\documentclass[letterpaper, 10pt, conference]{ieeeconf}      

\IEEEoverridecommandlockouts 
\usepackage{cite}
\usepackage{amsmath,amssymb,amsfonts}
\usepackage{graphicx}
\usepackage{textcomp}
\usepackage{amsmath}

\usepackage[table,xcdraw]{xcolor}
\usepackage{authblk}
\let\labelindent\relax
\usepackage{enumitem}
\usepackage{hyperref}
\usepackage{breqn}
\usepackage{algpseudocode}
\usepackage[linesnumbered,ruled,lined]{algorithm2e}
\usepackage{subfigure}
\usepackage{tcolorbox}
\usepackage{array}
\usepackage{makecell}

\allowdisplaybreaks

\begin{document} 

\let\ACMmaketitle=\maketitle

\title{ Hilbert Space Embedding-based Trajectory Optimization for Multi-Modal Uncertain Obstacle Trajectory Prediction   }
\author{Basant Sharma, Aditya Sharma, K.Madhava Krishna, Arun Kumar Singh\thanks{Aditya and Madhava Krishna are with IIIT-Hyderabad. Basant and Arun are with the University of Tartu. This research was in part supported by financed by European Social Fund via ICT program measure, grant PSG753 from Estonian Research Council and collaboration project LLTAT21278 with Bolt Technologies
Emails: aks1812@gmail.com. Code: \url{https://github.com/arunkumar-singh/RKHS_Stochastic_Traj_Opt.git}} 
}

\maketitle


\begin{abstract}
Safe autonomous driving critically depends on how well the ego-vehicle can predict the trajectories of neighboring vehicles. To this end, several trajectory prediction algorithms have been presented in the existing literature. Many of these approaches output a multi-modal distribution of obstacle trajectories instead of a single deterministic prediction to account for the underlying uncertainty. However, existing planners cannot handle the multi-modality based on just sample-level information of the predictions. With this motivation, this paper proposes a trajectory optimizer that can leverage the distributional aspects of the prediction in a computationally tractable and sample-efficient manner. Our optimizer can work with arbitrarily complex distributions and thus can be used with output distribution represented as a deep neural network. The core of our approach is built on embedding distribution in Reproducing Kernel Hilbert Space (RKHS), which we leverage in two ways. First, we propose an RKHS embedding approach to select probable samples from the obstacle trajectory distribution. Second, we rephrase chance-constrained optimization as distribution matching in RKHS and propose a novel sampling-based optimizer for its solution. We validate our approach with hand-crafted and neural network-based predictors trained on real-world datasets and show improvement over the existing stochastic optimization approaches in safety metrics.
 
\end{abstract}

\section{Introduction}
\begin{figure}
    \centering
    \includegraphics[scale=0.40]{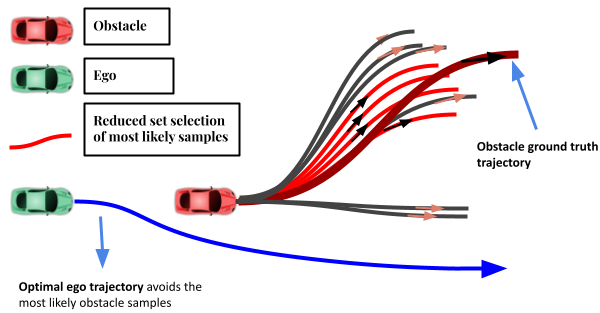}
    \caption{Figure shows a scenario where an obstacle has multiple intents (lane-change Vs lane-following), each associated with a trajectory distribution. However, both intents have wildly different probabilities. In this particular example, the probability of lane change is higher. For safe navigation, the ego-vehicle needs to consider this multi-modal nature of obstacle trajectories while planning its own motions. Our proposed approach estimates the more likely samples (the reduced set) from a set of obstacle trajectories sampled from a black-box distribution. This allows us to plan probabilistically safe motions while appropriately discriminating the low and high-probability obstacle maneuvers.  }
    \label{fig:teaser}
    \vspace{-0.7cm}
\end{figure}

Safety, or more precisely, collision avoidance, is a fundamental requirement in any autonomous driving system. It requires predicting how the world around the ego vehicle will evolve. As a result, trajectory prediction has become an extensively studied problem in the autonomous driving community. Our proposed work is focused on developing trajectory planners that can leverage the outputs of the current trajectory predictors in the best possible manner. To this end, we are inspired by a class of recent approaches that outputs a distribution of trajectories for the neighbouring vehicles (obstacles) instead of a single deterministic prediction. Works like \cite{ivanovic2019trajectron}, \cite{gupta2018social}, and \cite{lee2017desire} are a few popular algorithms in this regard. The distributional aspect of the trajectory prediction is crucial to capture the underlying uncertainty stemming from sensors or the unknown/unobserved intentions of the neighboring vehicles. For example, uncertainty in the intent can create complex multi-modal predictions (see Fig.\ref{fig:teaser}).

In this paper, we adopt a stochastic trajectory optimization perspective for motion planning of ego vehicles under uncertain obstacle trajectory prediction. There are two core challenges in this context. First, the analytical form of prediction distribution may be intractable or unknown. For example, works like \cite{ivanovic2019trajectron} characterise the output distribution through a deep generative model, drastically different from the Gaussian form \cite{zhu2019chance}. Thus, the convex approaches proposed in existing literature become unsuitable. Second, suppose we restrict our access to only samples drawn from the obstacle trajectory distribution. In that case, the optimizer should be able to compute high-probability collision avoidance maneuvers while considering only a handful of samples.

This work addresses both the above-mentioned challenges. Our proposed optimizer works with only sample-level information and thus is agnostic to the underlying distribution of the obstacle trajectories. We achieve this by building on the concept of embedding distribution into Reproducing Kernel Hilbert Space \cite{simon2016consistent}. In particular, given only drawn samples, we can represent the underlying distribution as a point in RKHS. This, in turn, opens up different possibilities. For example, it becomes straightforward to compute the difference between two distributions by embedding both in RKHS and computing the so-called Maximum Mean Discrepancy (MMD) measure \cite{simon2016consistent}. Our optimizer brings in two core innovations based on RKHS embedding and MMD for stochastic optimization. These are summarized below along with the associated benefits.

\subsubsection*{Algorithmic Contribution} 
\begin{itemize}
    
    \item We present a novel approach for selecting a subset of the most important/probable samples from the obstacle trajectory distribution. This subset is often referred to as the \emph{reduced-set} and lies at the very heart of our sample efficiency.  
    \item We reformulate stochastic optimization as a distribution-matching problem. The cost term is defined by the MMD measure and is conditioned on the trajectories of the ego-vehicle. 
    \item We present a custom sampling-based optimization for solving the distribution matching problem. We leverage a low-dimensional encoding of the trajectory sampling process and the introduction of a projection optimization to aid in constraint satisfaction.
\end{itemize}


\subsubsection*{State-of-the-art Performance}
\begin{itemize}
    \item We show that our one-shot \emph{reduced-set} selection method performs equal or better than a (near) exhaustive search for possible reduced sets.
    
    \item We perform extensive validation of our approach on multi-modal trajectory distribution from synthetic and real-world datasets. We show that our approach can handle uncertainty in manoeuvres like lane-change that stem from a complex mixture of discrete and continuous probability distributions. 

    \item We outperform recent work \cite{de2021scenario} in safety metrics when dealing with highly multi-modal trajectory distribution.
\end{itemize}

\section{Problem Formulation, Preliminaries and Related Works}
\noindent \subsubsection*{ Symbols and Notations} Scalars will be represented by normal-font small-case letters, while bold-faced variants will represent vectors. We will use upper-case bold fonts to represent matrices. Symbols $t$ and $T$ will represent the time-stamp and transpose operators, respectively. We will use $p_{(.)}$ to denote the probability density function of a random variable $(.)$. 

\subsection{Motion Planning in Frenet-Frame}
\noindent We formulate motion planning of the ego-vehicle in the road-aligned reference known as the Frenet frame. In this setting, the curved roads can be treated as ones with straight-line geometry. In other words, the longitudinal and lateral motions of the ego-vehicle are always with the $X$ and $Y$ axes of the Frenet-frame respectively.

\subsection{Feature Map and Kernel Function}
\noindent A feature map $\phi$ maps a feature $\textbf{z}$ to the Hilbert Space $\mathcal{H}$ as $\phi(\textbf{z})$. A positive definite kernel function $k$ is related to the feature map through the so-called kernel trick $k(\textbf{z}, \textbf{z}^{\prime}) = \langle \phi(\textbf{z}), \phi(\textbf{z}^{\prime})\rangle$. In this paper, we use the Gaussian kernel since they can capture all the possible moments of the underlying distribution \cite{simon2016consistent}.



\begin{table}[!t]
\centering
\caption{Computation Time vs Number of Samples generated by Trajectron++ (seconds) (Mean/Min/Max)}
\label{trajectron_inferencing}
\scriptsize
\begin{tabular}{|l|l|l|l|l|l|l|l|}
\hline
Number of Samples & Mean/Min/Max \\ \hline
16 samples & 0.0745 / 0.0737 / 0.0764  \\ \hline
32 samples & 0.0743 / 0.0734 / 0.0781  \\ \hline
64 samples & 0.0748 / 0.0738 / 0.0776  \\ \hline
128 samples & 0.0803 / 0.0787 / 0.0847  \\ \hline
256 samples & 0.0964 / 0.0954 / 0.0989  \\ \hline
512 samples & 0.1299 / 0.1288 / 0.1493  \\ \hline
1024 samples & 0.2018 / 0.2004 / 0.2122  \\ \hline
\end{tabular}
\normalsize
\vspace{-0.5cm}
\end{table}

\subsection{Stochastic Optimization}
\noindent Let $(x[k], y[k])$, $(x_{o}[k], y_{o}[k])$ be the ego-vehicle and obstacle trajectory waypoint at time step $k$. The latter is supposed to be a random variable belonging to some unknown distribution $p_o$. We can formulate stochastic trajectory optimization for ego-vehicle in the following form, wherein $(.)^{(q)}$ represents the $q^{th}$ derivative of the variable. We use $P(.)$ to denote the probability of a random variable $(.)$.

\small
\begin{subequations}
\begin{align}
    \sum_k \ddot{x}[k]^2+\ddot{y}[k]^2+(\dot{x}[k]-v_{des})^2 \label{cost}\\
    (x^{(q)}[k_0], y^{(q)}[k_0], x^{(q)}[k_f], y^{(q)}[k_f]) = \textbf{b} \label{eq_constraints}\\
    \textbf{g}(x^{(q)}[k], y^{(q)}[k]) \leq 0, \forall k \label{ineq_constraints}\\
    P(f(x[k], y[k], x_{o}[k], y_{o}[k] )\leq 0) \geq \eta. \forall k \label{chance_constraints}
\end{align}
\end{subequations}
\normalsize
\small
\begin{align}
    f = -\frac{(x[k]-x_{o} [k])^2}{a^2}-\frac{(y[k]-y_{o}[k] )^2}{b^2}+1
    \label{coll_constraints}
\end{align}
\normalsize
\noindent The first term in the cost function \eqref{cost} minimizes the acceleration magnitude while the second term aims to drive the ego-vehicle forward at the desired speed at each time step. The equality constraints \eqref{eq_constraints} ensure boundary conditions on the $q^{th}$ derivative of positions. For example, we consider the $0^{th}, 1^{st}, 2^{nd}$ derivatives in our formulation. The inequality constraints \eqref{ineq_constraints} model lane, velocity, and acceleration bounds.  The inequalities \eqref{chance_constraints} are referred to as chance constraints. They are responsible for ensuring that the ego-vehicle trajectory avoids the obstacle trajectory distribution with some lower bound confidence $\eta$. To this end, $f(.)$ is a regular collision-avoidance constraint as shown in \eqref{coll_constraints}, wherein we have assumed that the ego-vehicle and obstacles are represented as axis-aligned ellipses with size $\frac{a}{2}, \frac{b}{2}$. Extension to more sophisticated models is straightforward. See Section \ref{shapes}.

Beyond simple cases where $p_o$ is Gaussian, optimizations of the form \eqref{cost}-\eqref{chance_constraints} are computationally intractable. In this work, we focus on the case where the form of $p_o$ is not known and we only have access to the samples drawn from it. With this setting in place, this paper's core problem can be summarized as follows.

\begin{tcolorbox}[colback=green!5!white,colframe=green!75!black]
\noindent \emph{Problem P: Let $\textbf{O}$ be the matrix containing trajectory samples drawn from $p_o$. Let $\overline{\textbf{O}}$ be a matrix that contains $m$ out of a total $n$ obstacle trajectory samples in the matrix $\textbf{O}$. We call $\overline{\textbf{O}}$, the reduced-set. In the limiting case, $m =n$. However in general, $m<<n$. Then :}
\emph{
\begin{itemize}
    \item P.1: How can the $m$ subset of samples be selected?
    \item P.2: How to reformulate the chance constraints \eqref{chance_constraints} such that stochastic optimization formulated with $m$ samples from $\overline{\textbf{O}}$ generalizes well to the unseen samples from $p_o$.
\end{itemize}}

\end{tcolorbox}
It is easy to deduce that both parts \emph{P.1}, and \emph{P.2} are coupled. For example, some reformulations of \eqref{chance_constraints} may make the choice of reduced-set less critical. Alternately, an informed choice of reduced-set could aid in the generalizability of even basic approximations of \eqref{chance_constraints}. In the next subsection, we discuss some existing approaches for solving \emph{P.1} and \emph{P.2}.


\subsection{Related Works}
\noindent \textbf{Chance Constraints Approximation:} We primarily focus on existing works that can work with sample-level descriptions of uncertainty. In this context, the most popular reformulation for chance constraints is the scenario approximation \cite{calafiore2006scenario}, \cite{calafiore2012robust}. Here \eqref{chance_constraints}  is replaced with deterministic scenario constraints of the form $f(;, x_{o, j}[k], y_{o, j}[k])\leq 0$, defined with $j^{th}$ sample of obstacle trajectory prediction. A naive implementation of scenario approximation can be overly conservative and display poor sample complexity. Both drawbacks, in turn, can be attributed to scenario approximations not considering the likelihood of an obstacle trajectory sample. In other words, collision avoidance constraints formulated with all the obstacle trajectory samples, irrespective of their likelihood, will be given equal importance during trajectory optimization. The concept of  Sample Average Approximation \cite{pagnoncelli2009sample} reduces the conservativeness as it allows for some of the constraints to be violated. In a recent series of works \cite{poonganam2020reactive}, \cite{gopalakrishnan2021solving}, \cite{harithas2022cco}, we motivated the relaxation of \eqref{chance_constraints} as costs and subsequently expressed in terms of MMD between samples of $f(;, x_{o,j}[k], y_{o, j}[k])$ and those drawn from a specific proposal distribution.



\noindent \textbf{Reduced Set Selection:} One approach for reducing the conservativeness of vanilla scenario approximation is to perform problem-specific rejection sampling \cite{campi2011sampling}. For example, \cite{zhu2020kernel} presents a method where the vanilla scenario approach is first solved for many samples. Subsequently, the obtained solution is used to identify scenarios that can be discarded without affecting the confidence $\eta$ in a significant way. Although effective, this approach is not suitable for a real-time planning application as it requires repeated optimization calls. An improvement has recently been presented in \cite{de2021scenario} that performs rejection sampling based on the problem geometry. To be precise,  \cite{de2021scenario} rejects samples of $f(;, x_{o, j}[k], y_{o, j}[k])$ that are far away from the boundary of the feasible set. This rejection sampling is conceptually simple but requires evaluation of the rejection criteria on many samples. In other words, many obstacle trajectory samples must be drawn from $p_o$. This, however, can be prohibitive if there is a non-negligible computational cost associated with sampling. For example, Table \ref{trajectron_inferencing} presents the computation time required to draw various samples from Trajectron++ for a scene with five vehicles on an RTX 3060 i7-laptop with 16GB RAM. As can be seen, for real-time planning, drawing more than 100 samples could be challenging. Other deep neural network-based trajectory predictors show similar inferencing time. The rejection sampling of \cite{de2021scenario} can also give erroneous results when dealing with highly multi-modal distribution, as we also show latter.

\noindent \textbf{Contribution Over Author's Prior Work:} Our current work extends \cite{harithas2022cco}, \cite{poonganam2020reactive} to handle multi-step predictions of dynamic obstacle trajectories. In \cite{poonganam2020reactive}, the reduced set was formed with random sub-sampling of $\textbf{O}$. In contrast, we present a well-grounded approach that leverages some of the quintessential properties of RKHS embedding.

In the next section, we present our main algorithmic results, an improved reduced selection method, and its use in a reformulation of \eqref{cost}-\eqref{chance_constraints} that can better handle multi-modal obstacle trajectory predictions.




\section{Main Algorithmic Results}
\subsection{Proposed Reduced-Set Selection} \label{red_set_section}

\noindent Let $\boldsymbol{\tau}_{o,j} = (\textbf{x}_{o, j}, \textbf{y}_{o, j})$ be the $j^{th}$ obstacle trajectory formed by stacking the waypoints at different time steps $k$. Furthermore, let each obstacle trajectory be i.i.d samples drawn from $p_o$.  Then, $\sum_{j=1}^{j = n}\frac{1}{n}\phi(\boldsymbol{\tau}_{o, j})$ for some feature map $\phi(.)$ represents the embedding of the obstacle trajectory distribution in RKHS \cite{simon2016consistent}. There are two key advantages of RKHS embedding. First, it can capture distribution level information with a much smaller sample size $n$. Second, we can change the weight of each sample from $\frac{1}{n}$ to some $\alpha_j$ and arrive at the new embedding  $\sum_{j=1}^{j = n}\alpha_j\phi(\boldsymbol{\tau}_{o, j})$ with minimal loss of information. The latter forms the backbone of our reduced-set selection. 

Imagine that in the process of re-weighting, some of the $\alpha_j$'s assume a much higher value than the rest. For example, any $10 \%$ of the samples of $\textbf{O}$ has a substantially larger magnitude than the remaining $90 \%$. We then keep those specific $10 \%$ samples and discard the rest to form our reduced-set $\overline{\textbf{O}}$. We formalize our idea through the following optimization problem, wherein $\boldsymbol{\alpha} = (\alpha_1, \alpha_2, \dots, \alpha_n)$

\vspace{-0.5cm}
\small
\begin{align}
    \min_{\boldsymbol{\alpha}} \overbrace{\left\Vert \sum_{j=1}^{j = n}\frac{1}{n}\phi(\boldsymbol{\tau}_{o, j})-\sum_{j=1}^{j = n}\alpha_j\phi(\boldsymbol{\tau}_{o, j})\right\Vert_2^2}^{MMD}-\beta \frac{\sum \vert\overline{\alpha }_j (\boldsymbol{\alpha})\vert}{\sum\vert \widetilde{ \alpha}_j(\boldsymbol{\alpha}) \vert},
    \label{red_set_opt_proposed}
\end{align}
\normalsize

\noindent where $\overline{\alpha }_j$ and $\widetilde{ \alpha}_j$ respectively represent the top $m$ and bottom $n-m$ elements of $\boldsymbol{\alpha}$ in terms of magnitude. Thus, $\overline{\alpha }_j$ and $\widetilde{ \alpha}_j$ are (non-analytical) function of  $\boldsymbol{\alpha}$ and the parentheses in the second term of \eqref{red_set_opt_proposed} signifies this dependency. Note that the top and bottom samples are not defined beforehand. In contrast, the pattern itself is the output of \eqref{red_set_opt_proposed}.

The first term in cost \eqref{red_set_opt_proposed}, called the Maximum Mean Discrepancy (MMD), ensures that the re-weighting leads to an embedding close to the original one. The second term ensures that the re-weighting process creates clear differentiation between the more probable samples and the rest by increasing the magnitude of $\overline{\alpha}_j$ and vice-versa for $\widetilde{\alpha}_j$. The constant $\beta$ balances the two cost terms.

From here on, we will refer $\overline{\boldsymbol{\tau}}_{o, j} = (\overline{\textbf{x}}_{o, j}, \overline{\textbf{y}}_{o, j} ), \forall j = 1,2, \dots m$ as samples from reduced-set $\overline{\textbf{O}}$. Consequently, $\overline{\alpha }_j$ will be the weights of these samples. As we show later, these weights explicitly feature in our reformulation of \eqref{chance_constraints}. Thus, we follow the practical suggestion from \cite{smola1998learning} (pp.554) and we refine the magnitude of $\overline{\alpha}_j$ through \eqref{red_set_opt_proposed_2} to further minimize the information loss due to the discarded samples. Note that \eqref{red_set_opt_proposed_2} is done over the just-formed reduced-set.

\vspace{-0.1cm}
\small
\begin{align}
    \min_{\overline{\alpha}_j} \overbrace{\left\Vert \sum_{j=1}^{j = n}\frac{1}{n}\phi(\boldsymbol{\tau}_{o, j})-\sum_{j=1}^{j = m}\overline{\alpha}_j\phi(\overline{\boldsymbol{\tau}}_{o, j})\right\Vert_2^2}^{MMD}
    \label{red_set_opt_proposed_2}
\end{align}
\normalsize

\subsubsection{Solving \eqref{red_set_opt_proposed}} The proposed optimization \eqref{red_set_opt_proposed} is very challenging, mainly because the second term does not have an analytical form. Thus, we use a sampling-based optimization called CEM \cite{botev2013cross} to minimize \eqref{red_set_opt_proposed}. To ensure fast computation, we parallelize the cost evaluation over GPUs. Furthermore, we leverage the so-called kernel trick to factor out and pre-store the parts that do not depend on $\alpha_j$. Specifically, the first term reduces to the following form 

\small
\begin{align}
    \frac{1}{2}\left\Vert \sum_{j=1}^{j = n}\frac{1}{n}\phi(\boldsymbol{\tau}_{o, j})-\sum_{j=1}^{j = n}\alpha_j\phi(\boldsymbol{\tau}_{o, j})\right\Vert_2^2 = \frac{1}{2}\boldsymbol{\alpha}^T\textbf{K}\boldsymbol{\alpha}+\frac{1}{n}\textbf{K}\textbf{1},
\end{align}
\normalsize

\noindent where $\textbf{K}$ is the Gram matrix with $\textbf{K}_{jl} = k(\textbf{p}_{o, i}^j; \textbf{p}_{o, i}^l)$ for some kernel function $k(.)$. $\textbf{1}$ represent a vector of ones. As can be seen, the computationally expensive construction of $\textbf{K}$ needs to be done only once.

\subsubsection{Sanity Check} 
\begin{figure*}[t!]
\centering
 \includegraphics[scale=0.165]{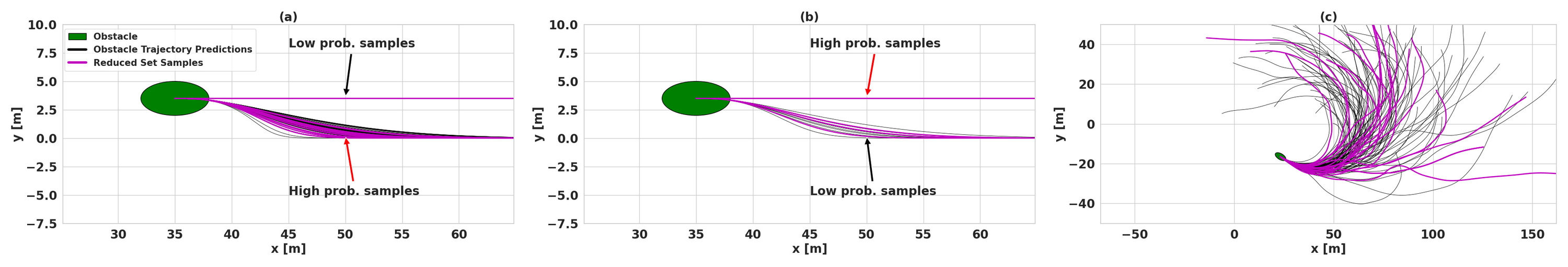}
\caption{Fig. (a) and (b) presents a multi-modal trajectory distribution that captures the uncertainty in the intent of lane change and how it will be executed. In Fig.(a) the vehicle is more likely to perform lane change while the situation is the opposite in Fig.(b). For safe autonomous driving, it is imperative that the planner be capable of handling these multi-modal uncertainties. In this paper, we build on RKHS embedding and propose a sample-efficient method. A key component of our approach is selecting probable samples (magenta) from just sample-level information. We call this the reduced set. In Fig.(a) and (b) the reduced set is primarily concentrated around the most probable manoeuvres. Fig. (c) shows the reduced-set selection for trajectories predicted from Trajectron++ on NuScenes dataset.}
\label{red_set_figures}
\vspace{-0.5cm}
\end{figure*}

We now present a synthetic example to showcase the inner working of our reduced-set selection. We consider a setting where a vehicle can perform lane-change or continue moving along its current lane. To model the uncertainty in the vehicle's motion, we sampled desired lateral offsets and velocity set-points from a discrete Binomial and Gaussian distribution, respectively. These are then passed to a trajectory optimizer to generate a distribution over the vehicle's motions. The results are summarized in Fig.\ref{red_set_figures}(a),(b). Due to the discrete nature of lateral offset distribution, we can precisely assign probabilities to each sample. It can be seen from Fig.\ref{red_set_figures}(a),(b) that reduced set samples are concentrated predominantly on the more probable manoeuvre. In Fig.\ref{red_set_figures}(a), the probability of lane change is higher, while in Fig.\ref{red_set_figures}(b), the vehicle is more likely to move along the current lane. For the sake of completeness, Fig.\ref{red_set_figures}(c) present the reduced-set selection for a scene using Trajectron++ prediction. Unfortunately, we can't assign probabilities to the samples since the underlying distribution is unknown \footnote{Trajectron++ and similar algorithms can predict a set of likely samples. But to the best of our knowledge, the exact probabilities of these samples are not known}. We provide more validation in Section \ref{red_set_validation_section}.

\subsection{Reformulating Chance Constraints}
\noindent In this subsection, we formulate a surrogate for $P(f(x[k], y[k], x_{o}[k], y_{o}[k] )\leq 0)$: an estimate of collision avoidance probability based on obstacle trajectory samples and conditioned on the ego-vehicle trajectory. To this end, we introduce the following relation.

\small
\begin{align}
    \overline{f} = \max(0, f)
    \label{f_bar}
\end{align}
\normalsize

\noindent Since $f$ is a random variable due to uncertain obstacle trajectories, so is $\overline{f}$. Let $p_{\overline{f}}$ be the probability distribution of $\overline{f}$. Although we don't know the exact characterization of $p_{\overline{f}}$, the definition of $\overline{f}$ guarantees that the entire mass of $p_{\overline{f}}$ will lie to the right of  $\overline{f} = 0$. Moreover, as $P(f(.)\leq 0)$ increases, $p_{\overline{f}}$ will tend to the Dirac-delta distribution $p_{\delta}$. Alternately, the difference between $p_{\overline{f}}$ and  $p_{\delta}$ can be used as the measure of the probability of collision avoidance \cite{harithas2022cco}.

As mentioned earlier, one way to measure the difference between two probability distributions is to embed them into RKHS and compute the MMD between the two. Let $\mu_{p_{\overline{f}}}$ and $\mu_{p_{\overline{f}}}$ be the RKHS embedding of $p_{\overline{f}}$ and $p_{\delta}$ respectively, then we use $l_{dist}$ as defined in \eqref{l_dist} as the measure of probability of collision avoidance. 

\vspace{-0.5cm}
\small
\begin{subequations}
\begin{align}
l_{dist} = \overbrace{\left\Vert \mu_{p_{\overline{f}}}-\mu_{p_{\delta}}\right\Vert_2^2}^{MMD}
\label{l_dist}\\
   \mu_{p_{\overline{f}}} = \sum_{j = 1}^{j = m} \overline{\alpha}_j\phi(\overline{f}(x[k], y[k], \overline{x}_{o, j}[k], \overline{y}_{o, j}[k] ) ) \label{mu_f_bar}\\
    \mu_{p_{\delta}} = \sum_{j=1}^{j = m} \phi(\delta_j) = \sum_{j=1}^{j = m} \phi(0)  \label{mu_delta}
\end{align}
\end{subequations}
\normalsize



\noindent Few important points about \eqref{l_dist}-\eqref{mu_delta} are in order

\begin{itemize}
    \item First, $l_{dist}$ is a deterministic scalar entity that depends explicitly on the trajectory waypoints of the ego-vehicle.
    \item Second, $\overline{x}_{o, j}[k], \overline{y}_{o, j}[k] $ are the obstacle trajectory samples from the reduced-set $\overline{\textbf{O}}$. Similarly, $\overline{\alpha}_j$ is the importance of these samples that were derived in Section \ref{red_set_section}.
    \item Finally, \eqref{mu_delta} leverages the fact that the samples from $p_{\delta}$ are all zero
\end{itemize}

\noindent We augment \eqref{l_dist} into the cost function \eqref{cost} in the following manner
\vspace{-0.4cm}
\small
\begin{align}
    c_{aug } = \sum_{k= k_0}^{k = k_f}\ddot{x}[k]^2+\ddot{y}[k]^2+(\dot{x}[k]-v_{des})^2+w\left\Vert \mu_{p_{\overline{f}}}-\mu_{p_{\delta}}\right\Vert_2^2
\label{c_aug}
\end{align}
\normalsize
\noindent where $w$ is used to trade-off the primary cost with MMD.

\subsection{Solution Process Using Sampling Based Optimization}
\noindent We minimize \eqref{c_aug} subject to \eqref{eq_constraints}-\eqref{ineq_constraints} through sampling-based optimization that combines features from CEM \cite{botev2013cross}, Model Predictive Path Integral (MPPI)\cite{williams2016aggressive} and its modern variants \cite{bhardwaj2022storm}. More importantly,  our approach leverages the problem structure of autonomous driving by sampling trajectories in the Frenet frame. It also incorporates a projection optimization to push the sampled trajectories towards feasible regions before evaluating their cost.

Our proposed optimizer is presented in Alg.\ref{algo_1}. Instead of directly sampling trajectories, we sample $\overline{n}_{cem}$ behavioural inputs $\textbf{d}_r$ such as desired lateral offsets and longitudinal velocity set-points from a Gaussian distribution (line 4). These are then fed into a Frenet space planner inspired by \cite{wei2014behavioral} in line 6, effectively mapping the distribution over behavioural inputs to that over trajectories. We present the details about the Frenet planner in Appendix \ref{appendix}. The obtained trajectories are then passed in line 8 to a projection optimization that pushes the trajectories towards a feasible region. Our projection problem is a special case of that proposed in \cite{masnavi2022visibility}, \cite{adajania2022multi} and thus can be easily parallelized over GPUs. In line 9, we evaluate the constraint residuals $c(.)$ over the projection outputs $(\widetilde{x}_{r}[k], \widetilde{y}_{r}[k])$. In line 10, we select top $n_{cem}(<\overline{n}_{cem})$ projection outputs with the least constraint residuals. We call this set the $ConstraintEliteSet$ and in line 11, we evaluate $c_{aug}(.)$ and $c(.)$ over this set. In line 13, we extract the top $n_{e}$ trajectory samples that led to the lowest combined cost and residuals. We call this the $EliteSet$. In line 14, we update the sampling distribution of behavioural inputs based on the cost sample collected in the $EliteSet$. The exact formula for mean and covariance update is presented in \eqref{mean_update}-\eqref{cov_update}. Herein, the constants $\gamma$ and $\eta$ are the temperature and learning rate respectively.

The notion of $EliteSet$ is brought from classic CEM, while the mean and covariance update follow from MPPI \cite{williams2016aggressive}, and \cite{bhardwaj2022storm}. An important feature of Alg.\ref{algo_1} is we have effectively encoded long horizon trajectories (e.g $k = 100$) with a low dimensional behavioural input vector $\textbf{d}$. This, in turn, improves the computational efficiency of our approach.


\small
\begin{subequations}
\begin{align}
    {^{l+1}}\boldsymbol{\mu}_d = (1-\eta){^{l}}\boldsymbol{\mu}_d+\eta\frac{\sum_{r=1}^{r=n_{e}} s_r\textbf{d}_r   }{\sum_{r=1}^{r=n_{e}} s_r}, \label{mean_update}\\
    {^{l+1}}\boldsymbol{\Sigma}_d = (1-\eta){^{l}}\boldsymbol{\Sigma}_d+\eta\frac{ \sum_{r=1}^{r=n_{e}} s_r(\textbf{d}_r-{^{l+1}}\boldsymbol{\mu}_{\textbf{d}})(\textbf{d}_r-{^{l+1}}\boldsymbol{\mu}_d)^T}   {\sum_{r=1}^{r=n_e} s_r} \label{cov_update}\\
    s_r = \exp{\frac{-1}{\gamma}(c_{aug}(\widetilde{x}_{r}[k], \widetilde{y}_{r}[k] )+c(\widetilde{x}_{r}[k], \widetilde{y}_{r}[k] )   }) \label{s_formula}
\end{align}
\end{subequations}
\normalsize

\subsection{Extension to Multiple Obstacles and Complex Shapes} \label{shapes}
\noindent Our approach presented in previous sections can be trivially extended to multiple obstacles. This is because both reduced-set optimization \eqref{red_set_opt_proposed} and the MMD surrogate \eqref{l_dist} can be independently constructed over all the obstacles. For extending to complex shapes, we recommend adopting the approach of covering the obstacle footprint with multiple circles. We can formulate MMD with respect to all the individual circles and stack them up together. This is the approach adopted in our implementation.

\vspace{-0.7cm}
\noindent 
 \begin{algorithm}[!h]
\caption{Sampling-Based MMD Augmented Trajectory Optimization}
\small
\label{algo_1}
\SetAlgoLined
$N$ = Maximum number of iterations\\
Initiate mean $^{l}\boldsymbol{\mu}_{d}, ^{l}\boldsymbol{\Sigma}_{d}$, at $l=0$ for sampling Frenet-Frame behavioural inputs\\
\For{$l=1, l \leq N, l++$}
{
Draw $\overline{n}_{cem}$ Samples $(\textbf{d}_1, \textbf{d}_2, \textbf{d}_r, ...., \textbf{d}_{\overline{n}_{cem}})$ from $\mathcal{N}(^{l}\boldsymbol{\mu}_d, ^{l}\boldsymbol{\Sigma}_d)$\\
 \vspace{0.1cm}
Initialize $CostList$ = []\\
 \vspace{0.1cm}

Query Frenet-planner for $\forall \textbf{d}_r$: $(x_{r }[k], y_{r}[k]) = \text{Frenet Planner}(\textbf{d}_r)$ \\
\text{Project to Constrained Set} \\
\small
\begin{align*}
    (\widetilde{x}_{r}[k], \widetilde{y}_{r}[k]) = \arg\min_{\widetilde{x}_{r}, \widetilde{y}_{r}} \frac{1}{2}\Vert \widetilde{x}_{r}[k]-{x}_{r}[k]\Vert_2^2\nonumber \\+\frac{1}{2}\Vert \widetilde{y}_{r}[k]-{y}_{r}[k]\Vert_2^2\\
     (\widetilde{x}_{r}^{(q)}[k_0], \widetilde{y}_{r}^{(q)}[k_0], \widetilde{x}_{r}^{(q)}[k_f], \widetilde{y}_{r}^{(q)}[k_f]) = \textbf{b} \\
    \textbf{g}(\widetilde{x}_{r}^{(q)}[k], \widetilde{y}_{r}^{(q)}[k]) \leq 0 \\
\end{align*}
\normalsize

Define constraint residuals: $c(\widetilde{x}_{r}[k], \widetilde{y}_{r}[k])$\\
$ConstraintEliteSet  \gets$ Select top $n_{cem}$   samples of $\textbf{d}_r, (x_{r }[k], y_{r}[k])$ with lowest constraint residual norm.\\
$cost \gets$ $c_{aug}(x_{r }[k], y_{r}[k])+r_q(x_{r }[k], y_{r}[k])$,  over $ConstraintEliteSet$ \\
 \vspace{0.1cm}
append ${cost}$ to $CostList$ \\
 \vspace{0.1cm}
          
$EliteSet  \gets$ Select top $n_{e}$ samples of ($\textbf{d}_r, x_{r }[k], y_{r}[k] $)  with lowest cost from $CostList$.\\

$({^{l+1}}\boldsymbol{\mu}_d, {^{l+1}}\boldsymbol{\Sigma}_d ) \gets$ Update distribution based on $EliteSet$

}
\normalsize
\Return{ Frenet parameter $\textbf{d}_r$ and  $(x_{r }[k], y_{r}[k])$ corresponding to lowest cost in the $EliteSet$}
\normalsize
\end{algorithm}

\begin{figure}[!t]
\centering
 \includegraphics[scale=0.21]{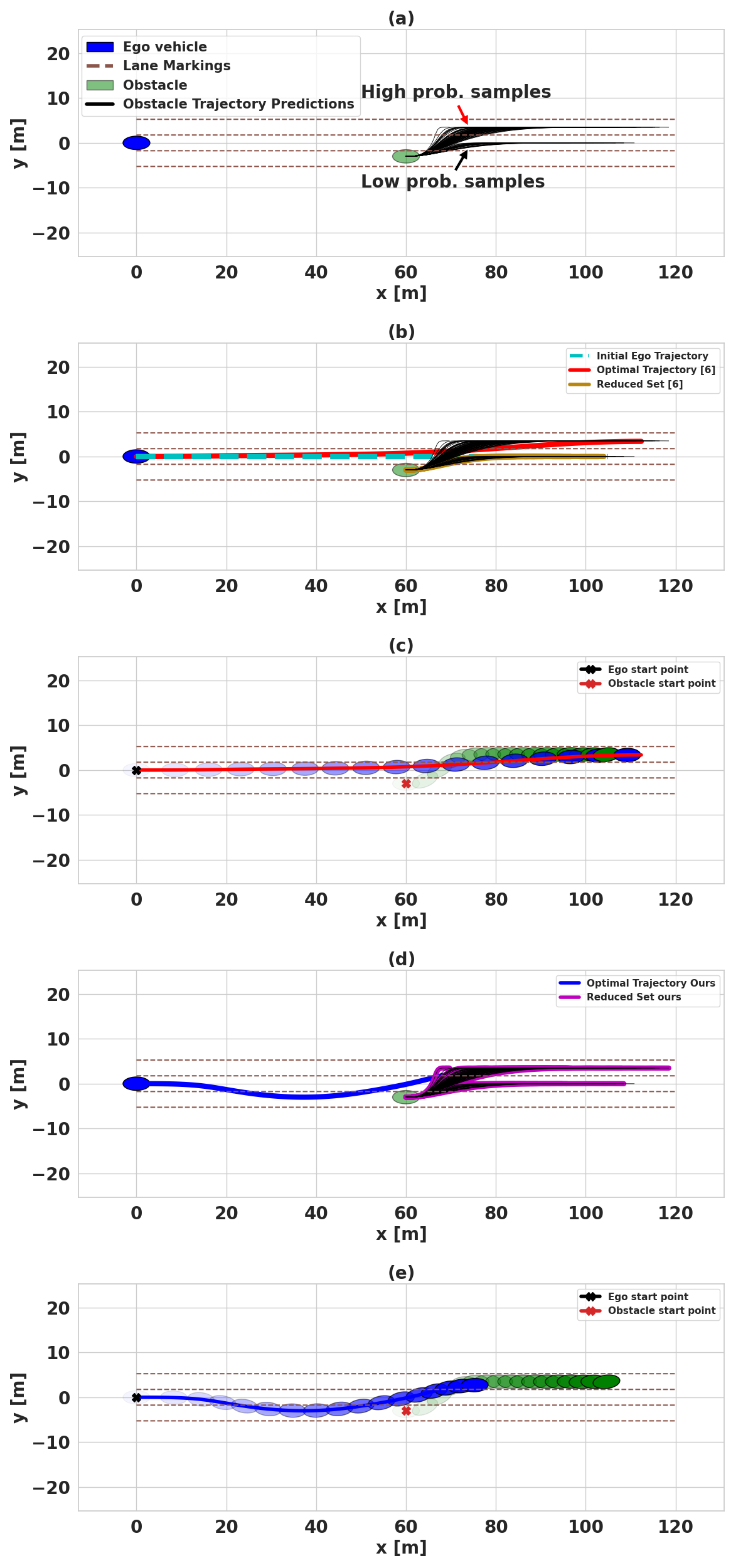}
\caption{Comparison of our approach and \cite{de2021scenario} on synthetic multi-modal dataset. Fig.(a) shows a scene with uncertainty in lane-change intent and its execution. Fig.(b) shows the reduced set selected following the criteria of \cite{de2021scenario} and the resulting optimal trajectory. As can be seen, the reduced set captured only less probable samples which led to the collision with a novel sample from the validation set in Fig.(c). In contrast, our reduced set optimization \eqref{red_set_opt_proposed} selects samples primarily from the high probability trajectory. But, some samples from the low probable regions are also selected. As a result, our MMD-based collision surrogate \eqref{l_dist} gets a correct estimate of the risk. Fig.(e) shows the trajectory computed through our \textbf{MMD-Opt} avoids the novel sample that \cite{de2021scenario} collided with. In Fig.(c)-(e)  }
\label{qual_result_1}
\vspace{-0.5cm}
\end{figure}

\section{Validation and Benchmarking}
This section aims to answer the following questions

\begin{itemize}
    \item \textbf{Q.1} How well our approach works with the commonly occurring multi-modal prediction of obstacle trajectories vis-a-vis the state-of-the-art (SOTA)?

    \item \textbf{Q.2} How well our reduced-set selection captures the probable samples from obstacle trajectories?
\end{itemize}

\subsection{Implementation Details}
\noindent We implemented our reduced-set selection optimization \eqref{red_set_opt_proposed} and Alg.\ref{algo_1} in Python using JAX as our GPU-accelerated numerical algebra library. The hyperparameters of Alg.\ref{algo_1} were $ \overline{n}_{cem} = 1000, n_{cem}=150, n_e = 50 $. We used $\gamma = 0.9$ and $\eta = 0.6$ in \eqref{mean_update}-\eqref{cov_update}. We used a Gaussian kernel with a bandwidth of 30 in \eqref{red_set_opt_proposed}, \eqref{l_dist}.

\subsubsection{Baselines} We compare our approach based on optimal reduced set selection and Alg.\ref{algo_1}, henceforth referred to as \textbf{MMD-Opt} with the scenario Approximation of \cite{de2021scenario}. This baseline augments a vanilla scenario approach with a reduced set strategy. Essentially it identifies obstacle trajectories that lead to  $f(x[k], y[k], x_{o, j}[k], y_{o, j}[k])$ being zero for some given (initial guess) ego vehicle trajectory.
The chosen obstacle trajectory (reduced set) samples are then used in deterministic trajectory optimization. For a fair comparison, we use our sampling-based optimizer \ref{algo_1} to plan with the reduced set samples. We just replace our MMD cost with a deterministic collision cost.


\subsubsection{Benchmarks and Metrics} To evaluate \textbf{Q.1}, we used two types of datasets. An example of the first kind is shown in Fig.\ref{qual_result_1} where we hand-crafted a scene with uncertainty in the lane-change maneuver of the obstacle. That is, the obstacle can shift to different lanes in front of the ego-vehicle or can continue moving along its current lane. To model this behavior, we sampled different lane offsets and forward velocity set-points from a discrete Binomial and a Gaussian distribution respectively. We then pass them onto a Frenet frame planner \cite{williams2016aggressive}. We varied the probability assigned to each lateral offset and the initial position and velocity of the obstacle to construct 100 different scenes.

The second dataset we use is based on Trajectron++ predictions on NuScenes \cite{caesar2020nuscenes} (recall Fig.\ref{red_set_figures}). Trajectron++ is also capable of producing multi-modal predictions. However, the number of scenes with clear multi-modality is limited. We evaluated a total of 1300 scenes in this dataset. In each scene, we had access to a reference centerline and a designated ego vehicle. The obstacle was chosen as any other agent in the scene.

For both datasets, we sampled 100 trajectories in each scene and further choose 10 samples from them to form the reduced set. We recall that the reduced-set selection differs between ours and the baselines \cite{de2021scenario}. We further sampled 1000 novel samples of obstacle trajectories to validate the performance of our \textbf{MMD-Opt} and \cite{de2021scenario}. We call this set the \textbf{Validation Set}. 

We use the metric of a number of collision-free trajectories obtained on the \textbf{Validation Set} for comparing \textbf{MMD-Opt} and \cite{de2021scenario}.

\subsection{Benchmarking our Reduced Set Selection} \label{red_set_validation_section}
\begin{figure}[ht]
\begin{center}
\includegraphics[scale = 0.187]{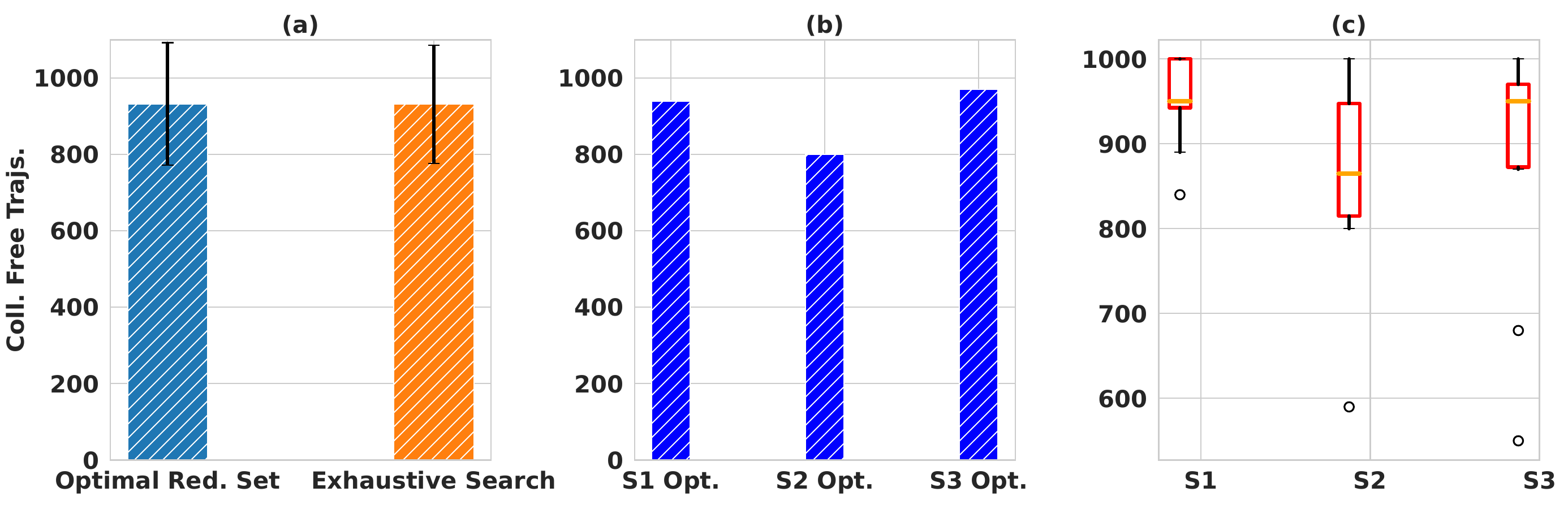}
\end{center}
  \caption{Fig. (a): Collision-free trajectories on the \textbf{Validation Set} obtained with our optimal reduced-set selection vis-a-vis through an exhaustive search over several random reduced set possibilities. As can be seen, our approach performs as well as an exhaustive search but in a tiny fraction of the computation time. Fig.(b)-(c) presents some specific scene examples ($S_1, S_2, S_3$) to showcase a fine-grained perspective. As can be seen, each different random choice for a reduced set offers wildly different performances. The average over all these performances is equal or lower than that obtained with the proposed optimal selection (blue bars in Fig.(b)).}
\vspace{-0.5cm}
\label{random_red_set_exhaustive}
\end{figure}

\noindent In this section, we evaluate the goodness of our reduced set selection. Since we don't have any ground truth to compare against, we take an indirect evidence-based approach. A good reduced set is one which leads to fewer collisions on the \textbf{Validation Set}. Our overall process was as follows. For each scene, we constructed several reduced sets by just randomly sub-sampling from the obstacle trajectory set. In other words, we performed some sort of (near) exhaustive search in the space of reduced sets. We then constructed the MMD \eqref{l_dist} using the $\overline{\alpha}_j$ associated with each of these reduced sets (recall \eqref{red_set_opt_proposed_2}) and subsequently solved Alg.\ref{algo_1}. The results are summarized in Fig.\ref{random_red_set_exhaustive}(a). It shows the mean and standard deviation of the number of collision-free trajectories achieved with our optimal reduced set selection and that with exhaustive search. As can be seen, our approach performs as well as an exhaustive search ($93.25\%$ Vs $93.13 \%$). It should be noted that in real-world scenarios, an exhaustive search is not possible and is done here solely for benchmarking. Fig.\ref{random_red_set_exhaustive}(b)-(c) presents a fine-grained perspective in three different scenes. As can be seen, different choices for the reduced set offer wildly different performances. Moreover, it is also not possible to know beforehand the performance of a random reduced set unless we have solved Alg.\ref{algo_1} with that choice. In contrast, our proposed solution based on minimization of \eqref{red_set_opt_proposed} offers a one-shot solution.

\subsection{Handling Multi-Modal Obstacle Trajectories}
\subsubsection{A Qualitative Result}
\noindent Let us assume that in the scene shown in  Fig.\ref{qual_result_1}, there is a $5\%$ probability of the obstacle merging into the lane of the ego-vehicle and a $95 \%$ percent probability of it choosing the adjacent lane. Interestingly as shown, the less probable obstacle trajectory samples are in direct conflict with the current trajectory of the ego vehicle. In contrast, the high probability maneuver takes the obstacle away from the ego vehicle. In other words, the less probable samples are at the boundary of the feasible set of collision avoidance constraints. Thus, if we apply the reduced set criteria from \cite{de2021scenario}, the  most conflicting but low probability samples (golden lines in Fig.\ref{qual_result_1}(b)) will be selected. Unfortunately, once fed to the optimizer, the resulting solution will lead the ego vehicle directly in conflict with more probable obstacle trajectory samples. The ensuing collision is documented in Fig.\ref{qual_result_1}(c).

Our approach operates in a strikingly different manner in this particular example and in general. First, our reduced set selection correctly identifies the high-probability samples (magenta Fig.\ref{qual_result_1}(d)). In fact, a small amount of reduced set samples are also chosen from the less probable ones.  Furthermore, our MMD optimization leads to the right set of manoeuvres for the ego-vehicle, validated for a novel random sample in Fig.\ref{qual_result_1}(e)

\subsubsection{Quantitative Validation} 
\begin{figure}[!t]
\centering
 \includegraphics[scale=0.26]{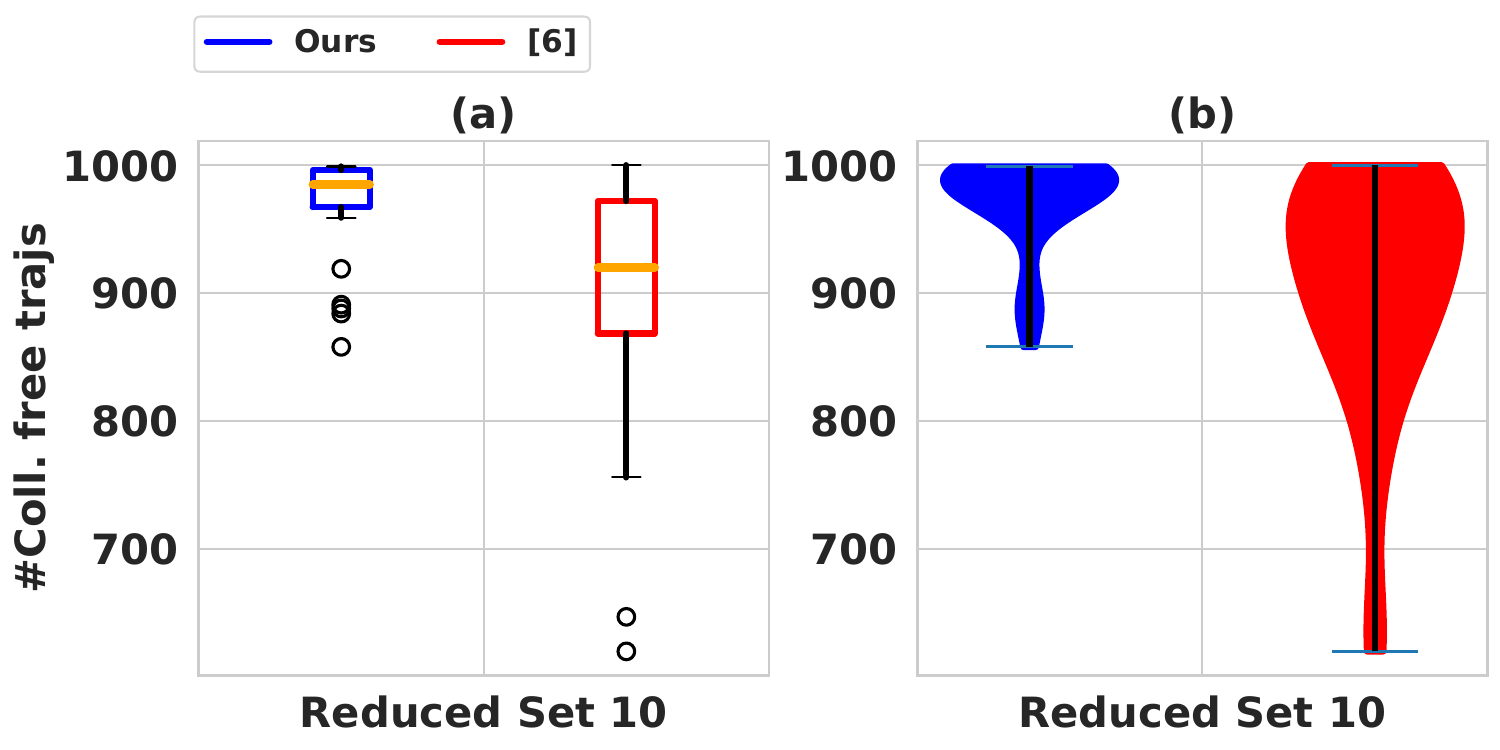}
\caption{Comparison of our approach and \cite{de2021scenario} on synthetic multi-modal dataset. We achieve more collision-free trajectories (out of 1000) more consistently than \cite{de2021scenario}.  }
\label{f_bar_plot_synthetic}
\vspace{-0.5cm}
\end{figure}

We constructed a total of 100 scenes similar to Fig.\ref{qual_result_1}(a). For each of these scenes, we evaluated the computed optimal trajectory (\textbf{MMD-Opt} and \cite{de2021scenario}) on the novel set of 1000 obstacle trajectory samples. The statistics of the number of collision-free trajectories observed across scenes are presented in Fig.\ref{f_bar_plot_synthetic}. The mean collision-free trajectory obtained by our approach was 972 as compared to 906 obtained by \cite{de2021scenario}. Thus, our \textbf{MMD-Opt} achieved an improvement of around $7\%$ on average. However, a deeper insight can be obtained by looking at the variance of the data. As shown in Fig.\ref{f_bar_plot_synthetic}, the numbers obtained by \textbf{MMD-Opt} is heavily concentrated in the region between $950-1000$. In fact, our lower quartile number ($967$) is almost equal to the upper quartile numbers ($972$) obtained by \cite{de2021scenario}. In other words, our worst-case performance is equal to the best-case performance of \cite{de2021scenario}. Furthermore, our \textbf{MMD-Opt}'s best case number is almost $25\%$ higher than the worst-case performance of \cite{de2021scenario} (neglecting the outliers).

\begin{figure}[!t]
\centering
 \includegraphics[scale=0.26]{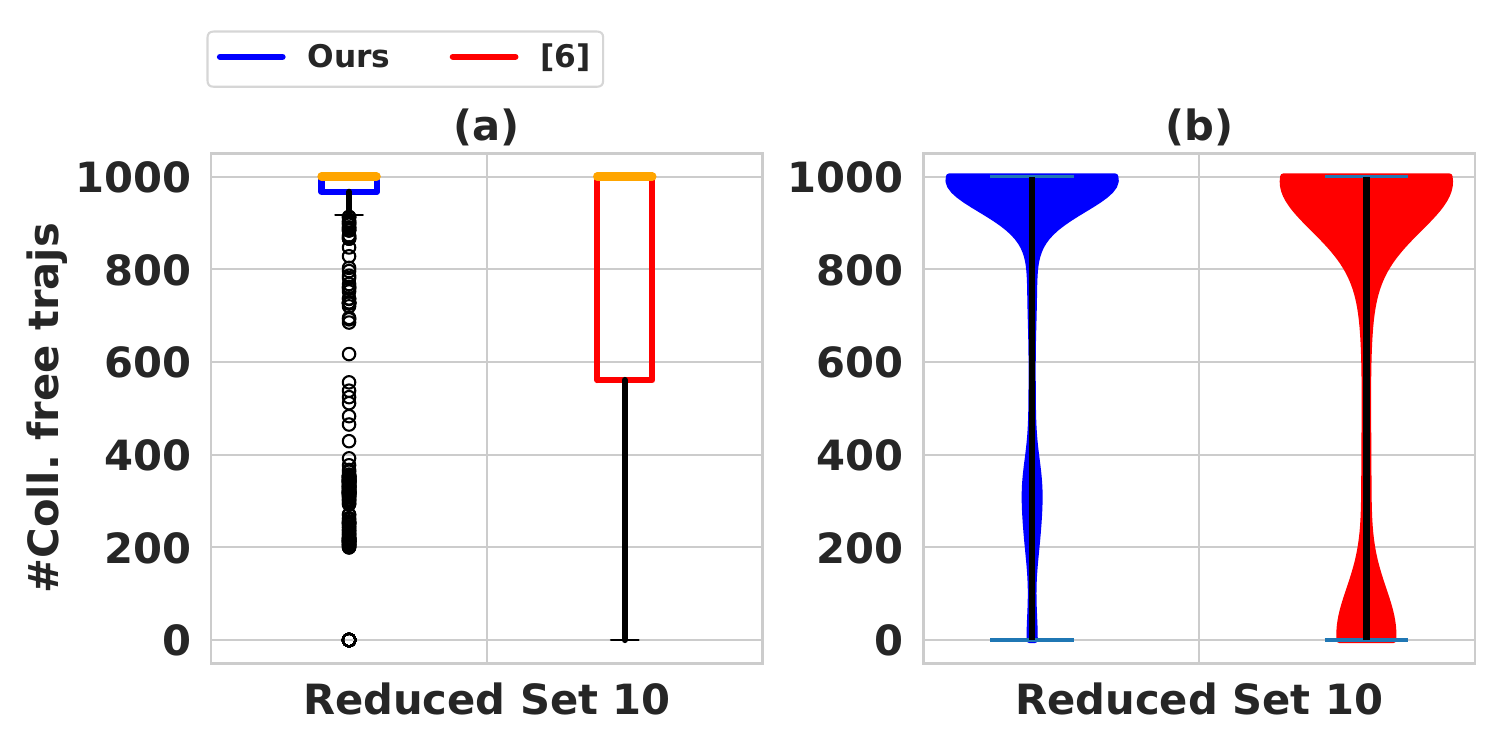}
\caption{Comparison of our approach and \cite{de2021scenario} on NuScenes dataset with Trajectron++ as the trajectory predictor. The trend in the number of collision-free trajectories is similar to that obtained in Fig.\ref{f_bar_plot_synthetic}: lower variance and heavy concentration around the top values. }
\label{f_bar_plot_trajectron}
\vspace{-0.4cm}
\end{figure}

Fig.\ref{f_bar_plot_trajectron} shows the statistics of collision-free trajectories on the NuScenes dataset with Trajectron++ as the multi-modal trajectory predictor. We see a similar trend as obtained before for the synthetic dataset: lower variance and heavy concentration around the upper bound. Moreover, across 1300 evaluated scenes, there was only one instance where our \textbf{MMD-Opt} obtained zero collision-free trajectories. In contrast, for \cite{de2021scenario}, the lower 25 $\%$ number is concentrated on zero.

\vspace{-0.3cm}
\subsection{Computation Time}
\noindent Table \ref{cem_timing} shows the computation time required for our reduced set optimization \eqref{red_set_opt_proposed} and Alg.\ref{algo_1} for different sample sizes on an RTX 3080 Laptop with 16 GB RAM. As shown, the timing of \eqref{red_set_opt_proposed} is independent of the number of samples we want to extract for the reduced set. However, Alg.\ref{algo_1}'s computation time increases depending on how many reduced set samples are used to form the MMD \eqref{l_dist}. Importantly, the timings are low enough to be considered real-time.

\begin{table}[!t]
\centering
\caption{Computation Time vs Number of Reduced set samples}
\label{cem_timing}
\scriptsize
\begin{tabular}{|c|c|c|}
\hline
Number of Reduced set samples & Computing \eqref{red_set_opt_proposed} & Alg. 1 \\ \hline
10 samples & 0.006 s & 0.02 s  \\ \hline
20 samples & 0.006 s & 0.025 s  \\ \hline
30 samples & 0.007 s & 0.035 s  \\ \hline
40 samples & 0.007 s & 0.05 s  \\ \hline
50 samples & 0.007 s & 0.07 s  \\ \hline
\end{tabular}
\normalsize
\vspace{-0.5cm}
\end{table}

\section{Conclusions and Future Work}
For the first time, we presented a trajectory optimizer that can efficiently handle multi-modal uncertainty including that on discrete intents (lane-change vs lane-keeping) of the dynamic obstacles. We showed how RKHS embedding can provide insights into more probable samples from obstacle trajectory distribution. This proves critical while estimating the likely maneuvers of the obstacles. Second, the same embedding leads to a surrogate for collision probability conditioned on the ego vehicle's trajectory. We proposed a sampling-based optimization for minimizing this collision surrogate while considering the typical kinematic constraints on the vehicle. 
We extensively compared against a very recent work \cite{de2021scenario} and showed that our approach outperforms it in safety metrics on both hand-crafted as well as real-world datasets.

Our work has certain limitations. The hyperparameters of the kernel function have a very strong effect on the overall performance of both reduced set selection as well as minimizing collision probability. One possible workaround is to use Bayesian optimization for tuning these parameters.  


\section{Appendix} \label{appendix}
\noindent Let $\textbf{d}= (y_d, v_d)$ be the lateral offset and desired velocity setpoints. Then Frenet planner \cite{wei2014behavioral} boils down to solving the following trajectory optimization.
\vspace{-0.3cm}
\small
\begin{subequations}
\begin{align}
  \min  \sum_k c_{s} +c_{l}+c_v\label{cost_frenet} \\
     (x^{(q)}[k_0], y^{(q)}[k_0], x^{(q)}[k_f], y^{(q)}[k_f]) = \textbf{b}  \label{boundary_cond_frenet}
\end{align}
\end{subequations}
\normalsize
\vspace{-0.5cm}
\small
\begin{subequations}
\begin{align}
    c_{s} (\ddot{x}[k], \ddot{y}[k]) = \ddot{x}[k]^2+\ddot{y}[k]^2\\
    c_{l}(\ddot{y}[k], \dot{y}[k]) = (\ddot{y}[k]-\kappa_p(y[k]-y_d)-\kappa_v\dot{y}[k])^2\\
    c_v(\dot{x}[k], \ddot{x}[k]) = (\ddot{x}[k]-\kappa_p(\dot{x}[k]-v_d))^2
\end{align}
\end{subequations}
\normalsize
The first term $c_s(.)$ in the cost function \eqref{cost_frenet} ensures smoothness in the planned trajectory by penalizing high accelerations at discrete time instants. The last two terms ($c_l(.), c_v(.)$) model the tracking of lateral offset ($y_{d}$) and forward velocity $(v_{d})$ set-points respectively with gain $(\kappa_p, \kappa_v)$. 


\vspace{-0.3cm}
\bibliography{ref_iros_mmd}
\bibliographystyle{IEEEtran}

\end{document}